\documentclass[10pt,A4paper,conference]{IEEEtran}
\IEEEoverridecommandlockouts
\usepackage{cite}
\usepackage{amsmath,amssymb,amsfonts}
\usepackage{graphicx}
\usepackage{textcomp}
\usepackage{xcolor}
\usepackage{tabularx}
\usepackage{color, colortbl}
\usepackage[misc]{ifsym}
\usepackage{xcolor}
\definecolor{lime}{rgb}{0.88,2,10}

\usepackage[left=.64in,
            right=.63in,
            top=.65in,
            bottom=0.93in]{geometry}
\renewcommand{\baselinestretch}{.86}
\usepackage{subcaption}
\usepackage{wrapfig}
\usepackage[linesnumbered,ruled,vlined]{algorithm2e}
\usepackage{xpatch}
\usepackage{url}
\usepackage{multirow}
\usepackage{multicol}
\usepackage{wrapfig}

\usepackage[export]{adjustbox}
\usepackage{nomencl}
\usepackage{siunitx}
\usepackage{blindtext}
\usepackage[T1]{fontenc}
\usepackage[spaces,hyphens]{xurl}
\usepackage{float}

\SetAlFnt{\small}

\SetKwComment{Comment}{$\triangleright$\ }{}

\usepackage[utf8]{inputenc}
\usepackage{enumitem}

\SetKwInput{kwInit}{Initialize}

\def\BibTeX{{\rm B\kern-.05em{\sc i\kern-.025em b}\kern-.08em
    T\kern-.1667em\lower.7ex\hbox{E}\kern-.125emX}}

\usepackage[numbers,sort&compress]{natbib}
\usepackage[font={small}]{caption} %it

\newcommand{\fref}[1]{Fig.~\ref{#1}}

\newcommand{\sref}[1]{Section~\ref{#1}}
\usepackage[misc]{ifsym}

\usepackage{tikz}
\newcommand*\circled[1]{\tikz[baseline=(char.base)]{
            \node[shape=circle,draw,inner sep=0.1pt] (char) {#1};}}
\newcounter{stepnum}
\newcommand{\steplabel}{\bfseries Phase \circled{\arabic{stepnum}}\stepcounter{stepnum}}

\usepackage[outline]{contour}
\contourlength{0.2pt}
\contournumber{15}

\newcommand\HUGE{\fontsize{22.3}{25}\selectfont}

\begin{document}
\title{\HUGE {CAV-AD: A Robust Framework for   Detection of Anomalous Data   and Malicious Sensors in CAV Networks}}

\author{\IEEEauthorblockN{
 Md Sazedur Rahman\IEEEauthorrefmark{2}, Mohamed Elmahallawy\IEEEauthorrefmark{2}, Sanjay Madria\IEEEauthorrefmark{2}, Samuel Frimpong\IEEEauthorrefmark{4}}  
      \IEEEauthorblockA{%
 \IEEEauthorrefmark{2}Computer Science Department, Missouri University of Science and Technology, Rolla, MO 65401, USA}
   \IEEEauthorblockA{%
 \IEEEauthorrefmark{4}Explosive \& Mining
Engineering Department, Missouri University of Science and Technology, Rolla, MO 65401, USA}
\thanks{$^*$This work was supported by a grant from CDC-NIOSH.} 
Emails:  {\{mrvfw, meqxk, madrias, frimpong@mst.edu\}}}
    
%\thanks{\IEEEauthorrefmark{2}Corresponding author. This work was supported by the National Science Foundation (NSF) under Grant No. 2008878.}  

\maketitle
\begin{abstract}

The adoption of connected and automated vehicles (CAVs) has sparked considerable interest across diverse industries, including public transportation, underground mining, and agriculture sectors. However, CAVs' reliance on sensor readings makes them vulnerable to significant threats. Manipulating these readings can compromise CAV network security, posing serious risks for malicious activities. Although several anomaly detection (AD) approaches for CAV networks are proposed, they often fail to: i) detect multiple anomalies in specific sensor(s) with high accuracy or F1 score, and ii) identify the specific sensor being attacked. In response, this paper proposes a novel framework tailored to CAV networks, called CAV-AD, for distinguishing abnormal readings amidst multiple anomaly data while identifying malicious sensors. Specifically, CAV-AD comprises two main components: i) A novel CNN model architecture called optimized omni-scale CNN (O-OS-CNN), which optimally selects the time scale by generating all possible kernel sizes for input time series data; ii) An amplification block to increase the values of anomaly readings, enhancing sensitivity for detecting anomalies. Not only that, but CAV-AD integrates the proposed O-OS-CNN with a Kalman filter to instantly identify the malicious sensors. We extensively train CAV-AD using real-world datasets containing both instant and constant attacks, evaluating its performance in detecting intrusions from multiple anomalies, which presents a more challenging scenario. Our results demonstrate that CAV-AD outperforms state-of-the-art methods, achieving an average accuracy of 98\% and an average F1 score of 89\%, while accurately identifying the malicious sensors.

\begin{IEEEkeywords} Anomaly detection, autonomous vehicles,  connected vehicles, sensors data.  \end{IEEEkeywords}
\end{abstract}
\section{Introduction}
The rapid advancements in sensor technology have made them suitable for various applications, including satellite communication (Satcom) \cite{10214895,10494442} and connected and autonomous vehicles (CAVs) \cite{ran2012dynamic,10103325}, enabling real-time communication with each other and the surrounding infrastructure. CAVs can operate without direct human input, sensing their environment, navigating roads, and autonomously making driving decisions. %This cohesive network enhances overall transportation management %\cite{dilek2023computer} and leads to the development of intelligent transportation systems (ITS). 
Due to their unmatched dependability and efficiency \cite{zhao2023effect}, CAVs have great potential for utilization in various other industrial sectors such as public transportation, underground mining \cite{sishi2020implementation,dong2020velocity}, and agriculture \cite{thomasson2018review}. %For example, in agriculture, CAVs can optimize farming methods and maximize crop yields while consuming fewer resources.
CAVs ability to operate in hazardous environments makes them valuable assets in industries like mining. Autonomous hauler trucks, for instance, can navigate rough terrain, tow large loads, and move commodities within mining sites with precision and efficiency. By substituting human operators for risky tasks, CAVs have the potential to increase industry safety and reduce the number of accidents and fatalities \cite{ferrein2023towards,ferrein2023controlling,de2024automation}.

%With the help of sophisticated sensing capabilities,
%When it comes to supply chain management and logistics, CAVs offer unmatched dependability and efficiency \cite{zhao2023effect}. 
%They have the potential to revolutionize transportation by improving safety, reducing traffic congestion, and enhancing mobility for people and goods.

To harness their significant advantages, CAVs depend an array of sensors, including cameras, radar, lidar, and GPS, in conjunction with advanced computing systems and AI algorithms \cite{lv2014traffic}. These components enable CAVs to perceive and interpret their surroundings accurately, plan optimal routes, and execute maneuvers safely and efficiently.

{\bf Challenges.}  While the utilization of CAV networks holds promise, their functionality and stability hinge on the robustness of their sensors against various attacks \cite{teh2020sensor,article,saeed2023review,gupta2023investigation,dong2023evaluating}. Typically, sensors in CAVs communicate over insecure channels, leaving them vulnerable to malicious entities seeking to compromise the entire CAV networks. Attackers may exploit to gain unauthorized access to sensor readings, monitor vehicle behavior, or even manipulate sensor readings for malicious purposes, potentially leading to catastrophic outcomes or fatal collisions \cite{zhang2021survey,ju2022survey,pandey4725372impact,el2020cybersecurity,wang2024anomaly}. As time-series datasets unfold sequentially, they are perceived as one-dimensional (1D), making them suitable for exploitation by a 1D-convolutional neural network (CNN), 1D-CNN, model network \cite{azizjon20201d} to extract temporal properties like traffic patterns and vehicle behaviors. An essential aspect of this process is determining the ideal receptive field size (RF), which represents the length of the input sequence scanned for features—a critical task for 1D-CNNs.
%\sout{Therefore, ensuring the security of CAVs' sensors is paramount to safeguarding the integrity and safety of the entire CAVs network.}

Several approaches have been developed to detect sensor integrity and intrusions in CAV networks. For example, Van et al. \cite{van2019real} proposed CNN-KF, which combines a CNN model with a Kalman filter (KF) for anomaly detection. %\sout{While CNNs automatically extract features and record temporal patterns, finding vulnerabilities in a dynamic system can be done effectively with the KF.} 
However, the performance of CNN-KF depends on the synchronous of both methods and if CNN fails to detect anomalies, KF may also struggle. Javed et al. \cite{9210741} introduced MSALSTM-CNN, a CNN based on long short-term memory (LSTM), which employs a multi-stage attention mechanism to improve anomaly identification rates. Despite its effectiveness, this approach cannot select optimal receptive field size for feature extraction and does not address the detection of malicious sensors. %\sout{ Recently, the authors of \cite{} proposed WKN-OC, which uses a wavelet kernel network and omni-scale convolutional blocks for anomaly detection. While effective in high-frequency signal processing, WKN-OC faces challenges in detecting a specific anomaly from a multiple anomaly type dataset. and does not address detection of malicious sensors.} 
%The artificial intelligence research community has also suggested various approaches \cite{wang2020real,lee2017attack}.

These methods, however, share common limitations: they failed i) to detect a specific anomaly type from multiple anomaly data, and ii) to identify the specific sensor(s) targeted in an attack.

{\bf Contributions.}In response to the challenges and limitations outlined above, this paper proposes a robust framework called CAV-AD for detecting anomalies in CAV networks, particularly those stemming from instant or constant attacks that pose significant risks and are more likely to occur in real-life scenarios. CAV-AD innovates through two key components: i) a novel model architecture named O-OS-CNN (\textbf{O}ptimized-\textbf{O}mni \textbf{S}cale {\bf CNN}), which is designed to select the optimal RF size for 1D-CNNs, thereby improving the detection performance in time series data; ii) an amplification block that enhances the sensitivity for detecting anomalies by increasing the values of anomaly readings. Additionally, CAV-AD incorporates a KF to dynamically identify malicious sensors.

In summary, this paper makes the following contributions:

\begin{itemize}

\item We propose CAV-AD, a framework designed to dynamically detect anomalies in CAV networks. Unlike existing approaches, CAV-AD succeeds in achieving high F1 scores alongside accuracy, demonstrating its effectiveness in anomaly detection. Furthermore, to the best of our knowledge, CAV-AD stands as the first framework with the capability of identifying malicious sensors that have been subjected to various attack scenarios, including instant, constant, gradual, and bias attacks.

\item CAV-AD innovates through two main components: (i) {\em novel AD model architecture called O-OS-CNN}, which dynamically adjusts the RF size of CNN by generating all possible kernel sizes across the entire length of the time series sequence used to extract features, allowing feature extraction from entire time series sequence; (ii)  {\em an amplification block} to enhance AD performance by capturing sudden abrupt changes from normal readings. This new block amplifies anomalous sensor readings with an amplification factor, making the model more sensitive toward sudden changes in sensor reading.

\item CAV-AD also integrates {\em the KF into our proposed O-OS-CNN model} to instantly detect the sensor with anomalous readings. Specifically, the KF predicts the next reading of each sensor data, and if the predicted value closely matches the measured one, it indicates no malicious sensor. However, if the predicted value deviates from the actual one by a certain threshold, it identifies the sensor as malicious.

\item We extensively evaluate CAV-AD using a real-world dataset for CAV networks, safety pilot model deployment (SPMD) dataset \cite{bezzina2014safety}, which is commonly used in SOTA research \cite{van2019real,9210741}. Our findings demonstrate that CAV-AD achieves an average accuracy of 98\% and F1 score of 89\% for detecting instant and constant anomalies across three different sensors, outperforming the SOTA approaches by an order of magnitude.

\end{itemize}

%Related works
\section{Related Works} \label{sec:related works}

%Although the field of AD in CAV networks is still in its nascent stage, 
To address the difficulties in identifying abnormal sensor data in CAVs, a number of previous works have been proposed. For instance, Javed et al. \cite{9210741} introduced a multi-stage attention mechanism utilizing a LSTM-based CNN, named MSALSTM-CNN, to identify abnormal sensor readings in autonomous vehicles. They also employed a weight-adjusted fine-tuned ensemble method to demonstrate the efficacy of the MSALSTM-CNN technique. Another approach by the authors of \cite{he2023vehicle} leveraged an attention-enhanced temporal convolutional network (TCN) for anomaly detection in CAV networks. This attention-enhanced TCN overcomes challenges associated with artificial feature selection and improves detection performance by adaptively assigning higher weights to important feature channels through attention branches at each layer.

Addressing instant attacks in sensor reading detection in CAV networks, Wyk et al. in \cite{van2019real} merged CNN with KF by combining the strengths of both models to outperform individual CNN and KF models. Khan et al. \cite{khan2021enhanced} presented a two-stage AD framework to identify instant attacks. Their approach involved utilizing a state-based Bloom filter technique in the first stage to confirm the states of incoming data, followed by a bidirectional LSTM (Bi-LSTM) classifier based on deep learning in the second stage to detect cyberattacks from autonomous vehicles. Cherdo et al. \cite{cherdo2023unsupervised} introduced semi-supervised anomaly threshold optimization and unsupervised anomaly detection models employing various state-of-the-art methods, including CNN, LSTM, Gated Recurrent Unit (GRU), and anomaly likelihood, to effectively detect anomalous temporal patterns in real complex multivariate sensor time series.

While these existing works have demonstrated promising performance, none of them have evaluated their approaches on multiple types of anomalies or attempted to predict a single type of anomaly. Furthermore, identifying the sensors under attack is a crucial aspect of cybersecurity that has been overlooked in previous research. To overcome these limitations, we introduce a novel framework called CAV-AD, which excels in detecting multiple types of anomalies compared to recent works and achieves higher detection accuracy and F1 scores. Additionally, CAV-AD is capable of identifying the sensors that are under attack. Further details about CAV-AD are discussed in \sref{sec:methodology}.

\begin{figure}[!t]
\centering
\includegraphics[width=0.49\textwidth]{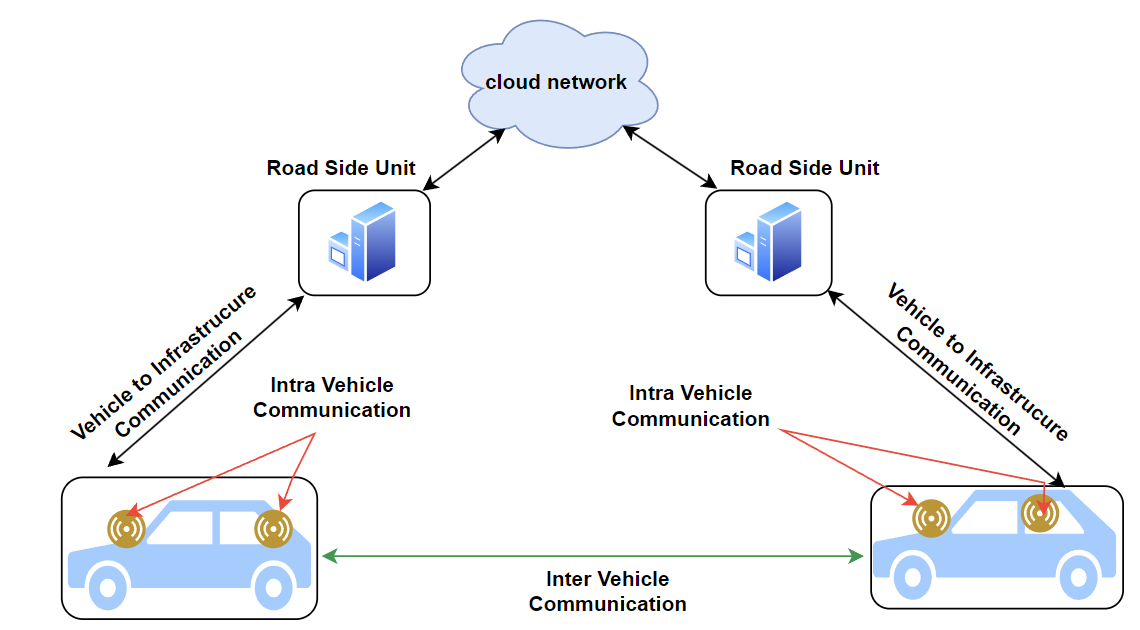}
\caption{Network Model.} 
    \label{fig:net-diag} 
\end{figure}

%By successfully implementing multiple beamforming in an automobile Muti-Input Multi-Output (MIMO) radar, Kapoor et al. \cite{kapoor2018detecting} have demonstrated a novel method to identify and prevent sensor spoofing attacks against automotive radars, which are crucial for assisted and autonomous driving. The method used by them is known as Spatio-Temporal Challenge-Response (STCR), which sends probe signals throughout time in a number of randomly chosen directions. When calculating the distance to the lead vehicle, STCR recognizes reflected signals from unreliable directions that cannot be confirmed based on the direction of arrival and eliminates them.

\section{Network And Threat Model}

In this section, we will provide an overview of our network model for the CAV network under consideration, along with the associated threats. 

\subsection{Network Model}

We consider a set of CAVs denoted by $\mathcal{C} = \{c_1, c_2, \ldots, c_N\}$, forming a network of $N$ autonomous vehicles. Each CAV $c_i$ is equipped with a set of sensors, represented by the set $\mathcal{S}_{C_i} = \{s_{1}, s_{2}, \ldots, s_{J}\}$. Each sensor $s_j$ within a vehicle $c_i$ can communicate with any other sensor on the CAV network. %for traffic control purposes. %The CAV network also comprises various other entities, including transport network infrastructure such as roadside units, transport control centers, and transport network assets, as well as networking devices like software-defined network switches and controllers.
Communication within our CAV network can be categorized into three classes: \textit{intra-vehicle communication}, where sensors within the same vehicle exchange information for decision-making, \textit{inter-vehicle communication}, where the computing management system of one vehicle communicates with others on the network, and {\em vehicle-to-infrastructure communication}, which is used to communicate with the network's main server (\fref{fig:net-diag} provides an illustration). On the one hand,  by collecting the information from the intra-vehicle sensors $\mathcal{S}_{c_i}$, a vehicle $c_i$ can optimize its driving path and make informed decisions, such as adjusting tire orientation and adapting lane position or speed. On the other hand, inter-vehicle communication is facilitated by establishing links with neighboring CAVs to reach the target vehicle. Additionally, each vehicle $c_i$ utilizes its onboard units for wireless connectivity to communicate with the main network's server through the network infrastructure. This communication typically involves passing through a hub on the main network, e.g., road side unit, to request information via vehicle-to-infrastructure communication.

\subsection{Threat Model}\label{sec:threat model}
%Despite the CAV network's promising potential, it faces some challenges. 

\begin{figure}[!t]
\centering
\includegraphics[width=0.49\textwidth]{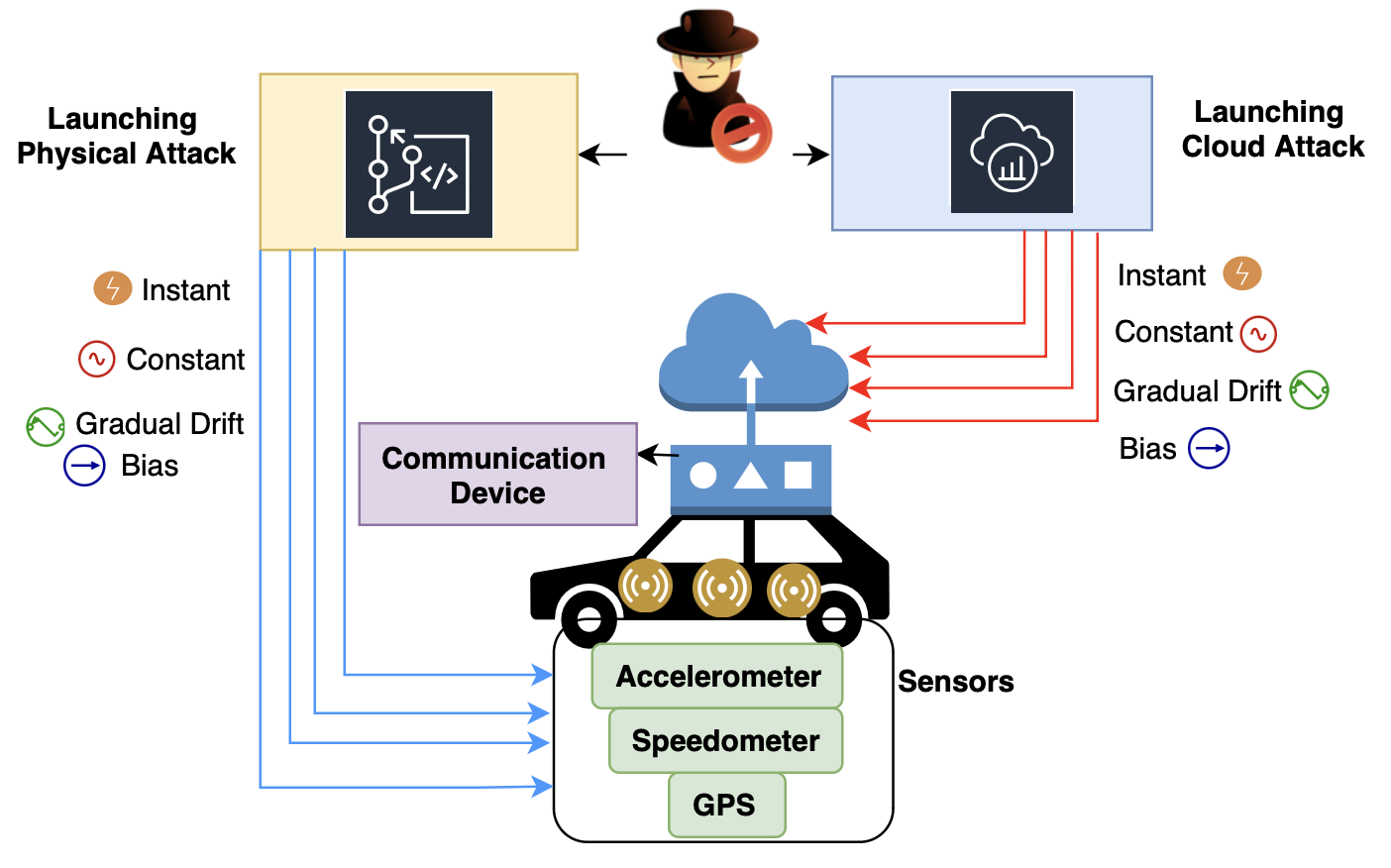}
\caption{Threat Model.} 
    \label{fig:sche-diag} 
\end{figure}

The dependence of CAVs on sensors, network infrastructure, and insecure communication channels renders them vulnerable to unforeseen malfunctions or attacks \cite{cui2019review,elmahallawy2023secure}. To this end, we consider four types of attacks/threats in our CAV network: \textit{instant attacks}, \textit{constant attacks}, {\em gradual drift attacks}, and {\em bias attacks} (see \fref{fig:sche-diag}). Below, we provide a brief description of each.

%There are two ways in which an attack can be carried out. The threat model is represented in a schematic diagram in 
%. The first method involves an attacker gaining remote access to the car's communication device and injecting malicious data into the sensors. The second method involves an attacker physically altering each sensor to replace the original readings.

%between CAVs takes place over potentially insecure channels, denoted by the set $\chi$. Additionally, communication among sensors within each vehicle also occurs, forming an intra-vehicle communication network.

\begin{enumerate} 
    \item {\bf Instant Attack.}
    An instant attack is defined as a quick, unexplained shift observed between the values of two consecutive CAV sensors. 
    
    \item {\bf Constant Attack.}
    This is an observation of temporarily constant for a time interval that deviates from typical sensor readings and is independent to the underlying physical phenomenon.
   \item \textbf{Gradual Drift Attack.} A small and gradual drift in observed data during a time period. 
  \item \textbf{Bias Attack.} A temporary constant offset from the sensor readings.
\end{enumerate}
In our threat model, we assume that only a single sensor will be a ``victim'' of any type of the above attacks. Thus, if anomalies come from multiple sensors simultaneously, it is easier to detect due to the inherent correlation among the sensors' readings \cite{gupta2008efficient}. This assumption aligns with SOTA research \cite{9210741, van2019real}.

\begin{figure*}[!t]
\includegraphics[width=\textwidth]{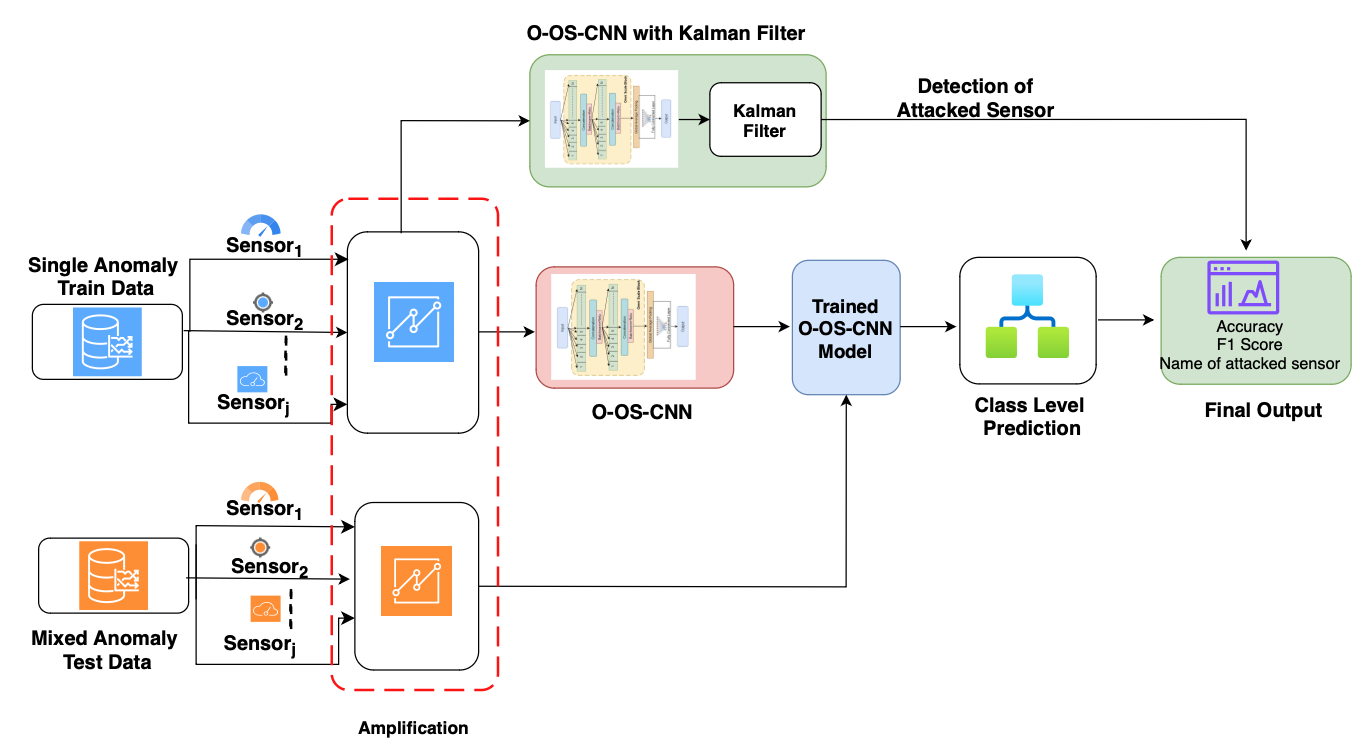}
\caption{An illustration of the CAV-AD framework.} 
    \label{fig:framework} 
\end{figure*}

\section{Proposed CAV-AD Framework} \label{sec:methodology}
In this section, we provide an overview of our proposed approach, the CAV-AD framework (illustrated in \fref{fig:framework}), designed to detect the threats outlined in \sref{sec:threat model} within CAV networks. The CAV-AD operates in three phases: (i) It utilizes the proposed \textit{amplification block} to enhance the sensitivity of AD, particularly for sensor readings with low-amplitude anomalies; (ii) It employs our novel \textit{O-OS-CNN model architecture}, which outperforms traditional CNN models in accurately identifying anomalous readings, achieving high accuracy and F1 scores for each sensor; (iii) It integrates the \textit{KF block} with our O-OS-CNN model to identify the sensor(s) under attack. Below, we provide a detailed description of how the CAV-AD framework operates:
\begin{algorithm}[!t]
\caption{CAV-AD's 3-phase process}\label{algorithm1}
\kwInit{Train Data $D_{\text{train}}$,  $D_{\text{test}}$}
\KwResult{List $S$ of malicious sensors, Accuracy, F1 Score}
 \Comment*[h]{{\bf Phase 1:} Data Amplification}\\{
  \SetKwProg{Fn}{Function}{}{}
    \Fn{CallAlgorithm 2}{
        \KwResult{Amplified $D_{\text{train}}$, Test Data $D_{\text{test}}$ }
    }
}
\Comment*[h]{{\bf Phase 2:} Train O-OS-CNN Model}\\{
\SetKwProg{Fn}{Function}{}{}
    \Fn{CallAlgorithm 3}{
        \KwResult{Accuracy, F1 Score}       
    }
}
\Comment*[h]{ {\bf Phase 3:} Detection Malicious Sensor(s)}\\
\SetKwProg{Fn}{Function}{}{}
    \Fn{CallAlgorithm 4}{
        \KwResult{List $S$ of malicious sensors }
        
    }
\textbf{Return} $S$, Accuracy, F1 Score
\end{algorithm}
\begin{algorithm}[!t]
    \SetAlgoLined
    \caption{CAV-AD Data Amplification Block}
    \label{algo:algo2}
    \KwIn{Anomaly Dataset: $D_{\text{train}}$ and $D_{\text{test}}$, Normal Dataset : $D_{\text{norm}}$ ,Threshold $\alpha$, Amplification Factor $\beta$}
    \KwOut{Amplified readings for sensors in $D_{\text{train}}$, $D_{\text{test}}$}%
    %Set a threshold $\alpha$\;
    %Set amplification factor $A$\\
    \For{each sensor $S_i$ in $D_{\text{train}}$, $D_{\text{test}}$}{
        \For{each epoch}{
            Find absolute difference, abs diff between current sensor reading and next reading\\
            \If{abs diff $> \alpha$ or abs diff ==0}{
                Amplify current reading by $\beta$\\
                Add to current reading \
            }
            \Else{
                Continue\
            }
        }
    }
    \textbf{Return} $D_{\text{train}}$, $D_{\text{test}}$
\end{algorithm}
\begin{itemize}[leftmargin=*, label={\steplabel}]
    \item  The upper amplification block of the CAV-AD framework on \fref{fig:framework} is fed with a dataset containing anomaly readings from a single sensor. Simultaneously, for testing purposes, the lower amplification block is fed with anomalies of four different attack types as outlined in \sref{sec:threat model} (see the bottom right of \fref{fig:framework}). Note that, both amplification blocks are designed to increase the magnitude of the anomaly readings, making the anomalies more distinguishable from normal readings.
 
    \item  The amplified data is subsequently utilized to train our proposed O-OS-CNN model. Concurrently, this amplified data is also used to train the O-OS-CNN model integrated with the KF, enabling the detection of the sensor from which the attack originates (see the middle of \fref{fig:framework}).

    \item For evaluation, the trained O-OS-CNN model will be fed with the amplified testing data to predict the class of the input data as either ``normal'' or ``anomaly''. The final output of the CAV-AD will include the accuracy and F1 score for anomaly sensor detection, along with the identification of the attacked sensor.
\end{itemize}

Algorithm \ref{algorithm1} outlines the three phases of the CAV-AD framework: the amplification block, the O-OS-CNN model architecture, and the O-OS-CNN empowered KF block. Each phase is executed by a separate algorithm, which will be discussed in detail in the following sections.

%In phase 1 amplification task is executed by calling Algorithm \ref{algo:algo2} which returns the amplified trained and test data. In phase 2, O-OS-CNN model is trained  using this amplified data by Algorithm \ref{algo:algo3}. Finally, to detect the list of attacked sensors, Algorithm 4 is executed.}

\subsection{CAV-AD Amplification Block}

We propose an amplification block to address the challenge of misidentifying or failing to detect anomalies with small amplitudes similar to normal readings, which can lead to reduced accuracy in AD \cite{he2023vehicle}. To achieve this, both the training and test datasets are passed through separate amplification blocks to enhance feature extraction, as illustrated in \fref{fig:framework}. Once CAV-AD detects a sudden deviation of a reading from its normal value, it triggers direct amplification.

%In scenarios involving vehicle anomaly detection, high-amplitude signals are prevalent  However,  traditional CNNs tend to align with low-frequency functions during training, making it challenging to learn anomalous signals with high-frequency components. This limitation educes the accuracy of anomaly identification  

To illustrate how our approach operates, we present Algorithm \ref{algo:algo2}, which detects such anomalies. We set a threshold value $\alpha$ to determine whether the absolute difference between the current reading and the next reading is sufficiently large to be amplified. Additionally, we define an amplification factor $\beta$ to amplify the sensor reading. In lines 1 through 12, the algorithm iterates over each reading of an individual sensor and then calculates the absolute difference between the current reading and the next reading. For instant anomaly ,if the difference exceeds the threshold $\alpha$ or for constant anomaly if the difference is zero, then it amplifies the current reading by a factor $\beta$, which is then added to the current reading.

The threshold value $\alpha$ is selected to be relatively large to prevent normal readings from being classified as abnormal, allowing the model to better distinguish these higher-amplitude anomaly readings. It is important to note that the value of $\alpha$ is not fixed and may vary depending on the statistical characteristics of the sensor data. For instance, the average absolute difference in one type of training set (instant) may differ from another (constant), influencing the choice of $\alpha$ based on the specific dataset.

%\orange{How did you choose the threshold and the amplification factor? We need to add away to explain that, otherwise, it looks vague.}\blue{added}\orange{Please any modification you will change should be highlighted in  blue color as I don't want to read again this section. I just need to check your modification.}

\subsection{O-OS-CNN Model Architecture}

This section presents our proposed O-OS-CNN model architecture designed for detecting anomaly data,i.e., sensor reading, with high accuracy and achieving a high F1 score. Unlike traditional 1D-CNNs used in time series classification tasks, which typically have fixed RF sizes that limit AD performance, O-OS-CNN tackles this challenge. Specifically, it dynamically adjusts its RF size based on the characteristics of the dataset, allowing it to select the optimal RF size for each training epoch. This capability is made possible by the inclusion of two-layered OS blocks within the O-OS-CNN architecture. These blocks enable O-OS-CNN to explore a range of kernel sizes, from 1 to the entire length of the time series data to extract features from the input in every possible length.

\begin{figure}[!t]
\includegraphics[width=0.5\textwidth]{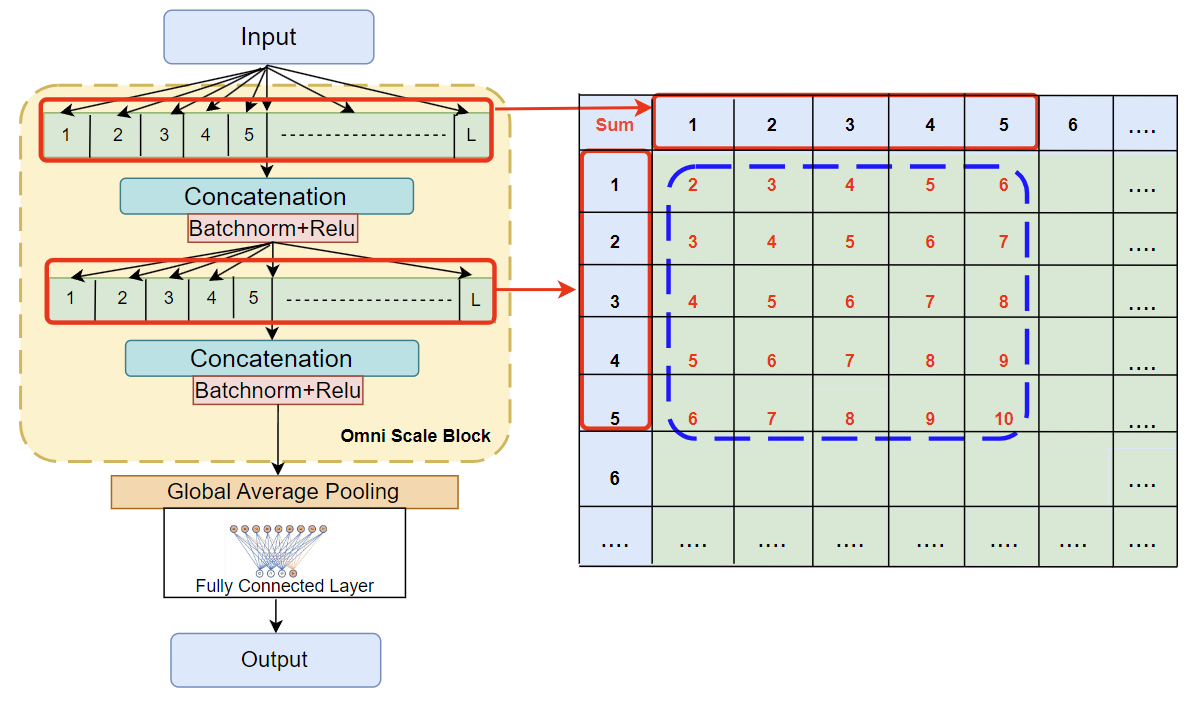}
\caption{An illustration of proposed O-OS-CNN model architecture.}
    \label{fig:architecture} 
\end{figure}

\fref{fig:architecture} illustrates the architecture of the O-OS-CNN model, which comprises two-layered OS blocks, while Algorithm \ref{algo:algo3} outlines the basic steps followed in this architecture. These blocks utilize a range of natural integers, from 1 to $L$, where $L$ is half the length of the time series data. Each kernel within the multi-kernel structure of these blocks performs the same input-to-padding convolution.

A similar OS block was proposed in \cite{tang2020omni}, which utilized only prime numbers in the initial OS block. However, this method has three layers, where the first two layers are comprised of prime numbers that can generate kernel sizes up to a certain range, depending on the length and type of dataset. Additionally, determining the smallest prime number that could yield optimal RF sizes within a specific range is challenging.

In contrast, our approach employs all possible numbers to explore every potential kernel size across the entire input length, thereby enhancing the range of feature extraction. For example, if the length of a time series dataset is 10, the value of $L$ in this instance would be 5.

In the OS block, the presence of multiple kernel sizes in each layer allows for the generation of various routes with different RF sizes from the input signal to the output feature. Following this, the output of the first layer within the OS block is concatenated, subjected to batch normalization, activated using the rectified linear unit (ReLU) function, and then forwarded to the second layer for further processing. This two-layered approach enhances the model's capability to extract meaningful features from the entire input data, leading to improved performance in AD. After the two-layered OS block, a fully connected layer is added for classification, and a global average pooling layer is employed for dimension reduction. On the right of \fref{fig:architecture}, we provide an example to illustrate the mathematical reasoning behind choosing a set of numbers from 1 to $L$, where each row and column represents the set of numbers for each layer of Omni Scale Block. The numbers inside the green grid denote the kernel sizes that could be generated by summing each pair of numbers from row and column. %The numbers inside the blue dotted box is an example of all possible kernel sizes of  length 2 to 10 made using integers from  1 to 5.

\begin{algorithm}[!t]
    \SetAlgoLined
    \caption{O-OS-CNN Model}
    \label{algo:algo3}
    
    \KwIn{Amplified Anomaly Dataset: $D_{\text{train}}$ and $D_{\text{test}}$}
    \KwOut{Accuracy, F1 Score}
    
    Set \( L \) as half the length of input data\\
    Initialize layer 1 of OS Block with kernel sizes \( 1, 2, \ldots, L \)\\
    Concatenate each kernel output to form feature maps: \( f_1, f_2, \ldots, f_n \)\\
    Apply batch normalization to each feature map: \( \hat{f}_1, \hat{f}_2, \ldots, \hat{f}_n \)\\
    Apply ReLU activation to each feature map: \( \text{ReLU}(\hat{f}_1), \text{ReLU}(\hat{f}_2), \ldots, \text{ReLU}(\hat{f}_n) \)\\
    Form final feature map  by concatenating the ReLU outputs: \( f_{\text{final}} = \text{Concat}(\text{ReLU}(\hat{f}_1), \text{ReLU}(\hat{f}_2), \ldots, \text{ReLU}(\hat{f}_n)) \)\\
    Apply global average pooling to \( f_{\text{final}} \) to obtain \( V \)\\
    Apply fully connected layer to \( V \) to obtain \( O \)\\
    Evaluate  \( O \)\ on  \( D_{\text{test}} \)\\
    \textbf{Return} Accuracy, F1 Score\\
\end{algorithm}

\subsection{O-OS-CNN Empowered Kalman Filter} \label{sec:detection}

We integrate our developed O-OS-CNN framework with a KF block to enhance the capability of our CAV-AD approach in detecting sensors that are under attack and exhibit anomaly reading (see the top of \fref{fig:framework}). This integration of the KF block introduces new functionality to the O-OS-CNN model, leveraging the effectiveness of the KF in estimating the state of dynamic systems in the presence of noisy sensor data.

One advantage of integrating the KF over alternative methods such as the Gaussian mixture model (GMM) is its superior performance in detecting attacked sensors. Our experimental comparisons between these two methods show that the KF outperforms the GMM due to its versatility in handling dynamic systems and high computing efficiency. Additionally, the KF is adept at estimating the state of a linear dynamic system from noisy measurements, operating in two main steps: prediction and update.

%Algorithm  \ref{algo:detection} is designed to serve this purpose using the strength of KF. 

%\textcolor{blue}{\textbf{Prediction Step:}
%\begin{align}
   % \hat{\mathbf{x}}^-_k &= A \hat{\mathbf{x}}_{k-1} + B \mathbf{u}_k \label{eq3}\\
   % P^-_k &= A P_{k-1} A^T + Q \label{eq4}
%\end{align}
%}

%\textcolor{blue}{\textbf{Update Step:}
%\begin{align}
  %  \mathbf{K}_k &= P^-_k H^T (H P^-_k H^T + R)^{-1} \label{eq5} \\
  %  \hat{\mathbf{x}}_k &= \hat{\mathbf{x}}^-_k + \mathbf{K}_k (\mathbf{z}_k - H \hat{\mathbf{x}}^-_k)\label{eq6}  \\
 %   P_k &= (I - \mathbf{K}_k H) P^-_k \label{eq7} 
%\end{align}
%}
%where:
%\begin{itemize}
   % \item $A$ is the state transition matrix,
   %\item $B$ is the control input matrix,
   % \item $H$ is the measurement matrix,
   % \item $\mathbf{w}_k$ is the process noise with covariance matrix $Q$,
   % \item $\mathbf{v}_k$ is the measurement noise with covariance matrix $R$,
  %  \item $\hat{\mathbf{x}}^-_k$ is the predicted state estimate,
   % \item $P^-_k$ is the predicted state covariance matrix,
   % \item $\hat{\mathbf{x}}_{k-1}$ is the previous state estimate,
   % \item $P_{k-1}$ is the previous state covariance matrix,
    %\item $\mathbf{K}_k$ is the Kalman gain,
    %\item $\hat{\mathbf{x}}_k$ is the updated state estimate,
   % \item $P_k$ is the updated state covariance matrix,
  %  \item $I$ is the identity matrix.
%\end{itemize}

Algorithm \ref{algo:detection} outlines the steps for utilizing the KF in each sensor to dynamically predict the next normal value based on the previous value. In line 1, we initialize a threshold $T$ to determine the gap between normal and abnormal values. Line 4 iterates over each sensor in the normal data, and line 6 initializes ${X}_0$ as the first value of a sensor. Then, in line 7, we predict the next normal value of that reading, $X_\text{pre}$, using KF, and store this predicted value in a list $\texttt{pre}\_\texttt{val}$.

From line 13, it iterates over the incoming anomaly data, with $X_\text{in}$ representing the current incoming reading. On line 16, the incoming value $X_\text{in}$ is stored in a list $\texttt{in\_val}$. Line 17 calculates the absolute difference between the incoming anomalous values and the normal predicted values by the KF. From lines 18 to 20, this absolute difference for each sensor is checked against the threshold to determine if the incoming value $X_\text{in}$ deviates significantly from the predicted value $X_\text{pre}$. The threshold $T$ is set to 2 because the normal sensor data (speed or acceleration) are typically much closer to each other than the abnormal ones. However, it can be adjusted based on statistical analysis of the dataset.

Ultimately, this algorithm generates a list of  {\em malicious sensors} that deviate significantly from the predicted values.

\begin{algorithm}[!t]
    \SetAlgoLined
    \caption{O-OS-CNN Empowered KF Model}
    \label{algo:detection}
    \KwIn{Normal Data $D_{\text{norm}}$, Anomaly Dataset $D_{\text{anom}}$}
    \KwOut{List $S$ of Attacked sensor $S_i$ if an anomaly is detected}
    \BlankLine
    Define the threshold, $T$ for absolute difference \\
    Initialize KF Parameters\\  
    Initialize an empty list $\texttt{pre\_val}$ for each sensor $S_i$\\
    \For{each row in $D_{\text{norm}}$}{
        \For{each sensor $S_i$}{
        ${X}_0\leftarrow$ \text{first sensor reading}\\
        $X_\text{pre}\leftarrow$ \text{Predicted value by KF} \\
        Update the KF  \\
        Append $X_\text{pre}$ to $\texttt{pre\_val}[S_i]$
        }
    }
     Initialize an empty list $\texttt{in\_val}$ for each sensor $S_i$ \\
    \For{each row in $D_{\text{anom}}$}{
        \For{each sensor $S_i$}
        {
        $X_\text{in} \leftarrow$ \text{sensor reading of }$S_i$\\
       $\texttt{in\_val}[S_i] \leftarrow$$X_\text{in}$\\
        $\texttt{abs\_diff} [S_i] = |\texttt{pre\_val}[S_i] - \texttt{in\_val}[S_i]|$\\
      \If{$\texttt{abs\_diff}~[S_i]  > T$}
      {
     Append sensor $S_i$ to list $S$
     }
     }   
    }
 \end{algorithm}

%Shorter Version of Algorithm 2
\begin{table}[!t]
\centering
\caption{Performance of CAV-AD for the Instant Anomaly.}
\label{table:instant}
\begin{tabular}{|c|c|c|c|c|}
\hline
Anomaly Magnitude             & Sensors & Acc  & Prec & F1   \\ \hline
Base value +  $1000\cdot \mathcal{N}(0, 0.01)$ & 1       & 99.3 & 83.1 & 83.4 \\ \cline{2-5} 
                              & 2       & 99.5 & 85.7 & 88.7 \\ \cline{2-5} 
                              & 3       & 99.5 & 83.0 & 88.0 \\ \hline
\end{tabular}
\end{table}

%Detection table for Constant Anomaly Magnitude
\begin{table}[!t]
\centering
\caption{Performance of CAV-AD for the Constant Anomaly.}
\label{table:constant}
\begin{tabular}{|c|c|c|c|c|}
\hline
Anomaly Magnitude                       & Sensors & Acc  & Prec & F1   \\ \hline
Base value +  $\mathcal{U}(0, 5), d=10s$ & 1       & 98.5 & 96.9 & 95.6 \\ \cline{2-5} 
                                        & 2       & 97.7 & 99.9 & 91.2 \\ \cline{2-5} 
                                        & 3       & 97.9 & 98.9 & 93.5 \\ \hline
\end{tabular}
\end{table}
\begin{figure*}[!t]
     \centering
     \begin{subfigure}[b]{0.325\textwidth}
         \centering
         \includegraphics[width=\textwidth]{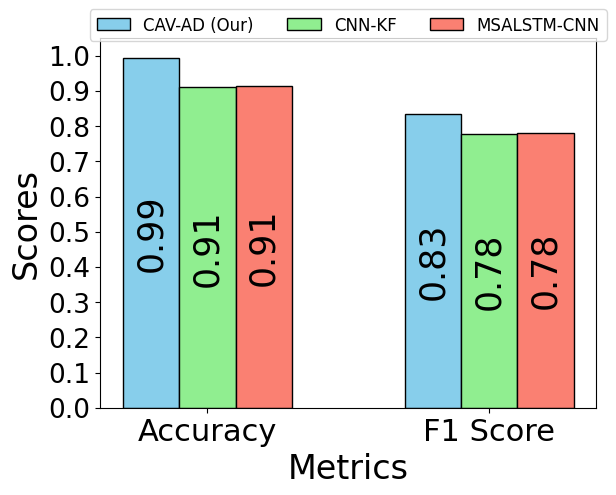}
         \caption{Sensor\#1.}\label{fig:Result_Instant_Sensor1}
     \end{subfigure}  
     \begin{subfigure}[b]{0.325\textwidth}
         \centering
         \includegraphics[width=\textwidth]{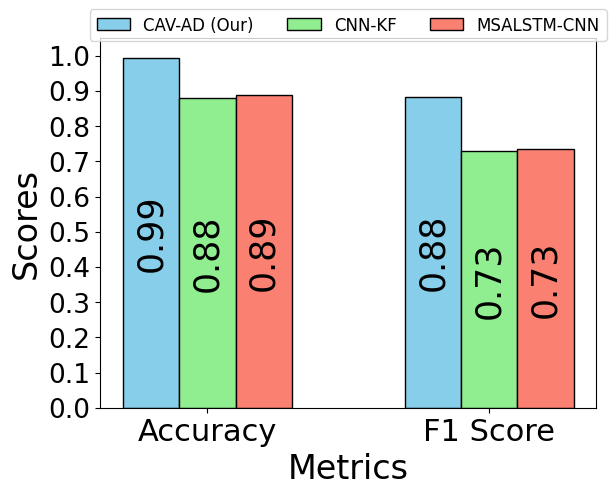}
         \caption{Sensor\#2.}\label{fig:Result_Instant_Sensor2}
     \end{subfigure}  
     \begin{subfigure}[b]{0.325\textwidth}
         \centering
         \includegraphics[width=\textwidth]{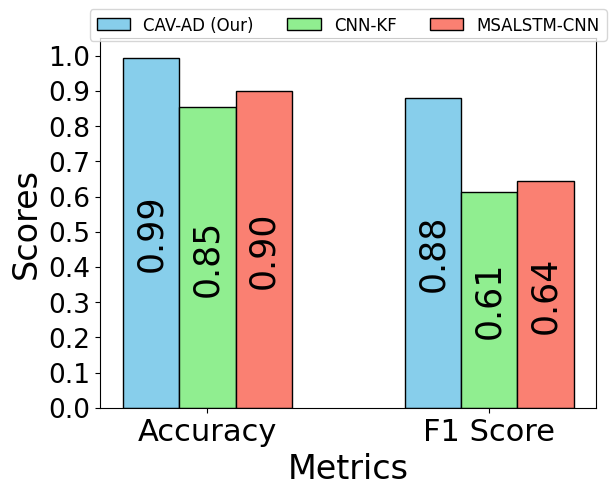}
         \caption{Sensor\#3.}\label{fig:Result_Instant_Sensor3}
     \end{subfigure}  
     \caption{Comparison of Accuracy and F1 score among CAV-AD, CNN-KF, and MSALSTM-CNN for instant anomaly detection.}
     \label{result:instant}
\end{figure*}

\begin{figure*}
     \centering
     \begin{subfigure}[b]{0.325\textwidth}
         \centering
         \includegraphics[width=\textwidth]{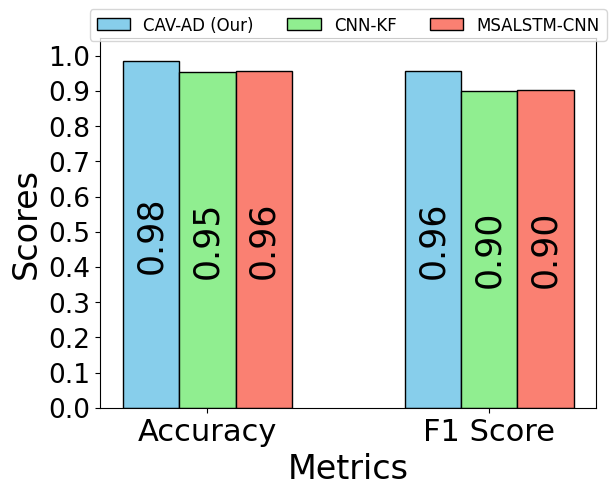}
         \caption{Sensor\#1.}\label{fig:Result_Constant_Sensor1}
     \end{subfigure}  
     \begin{subfigure}[b]{0.325\textwidth}
         \centering
         \includegraphics[width=\textwidth]{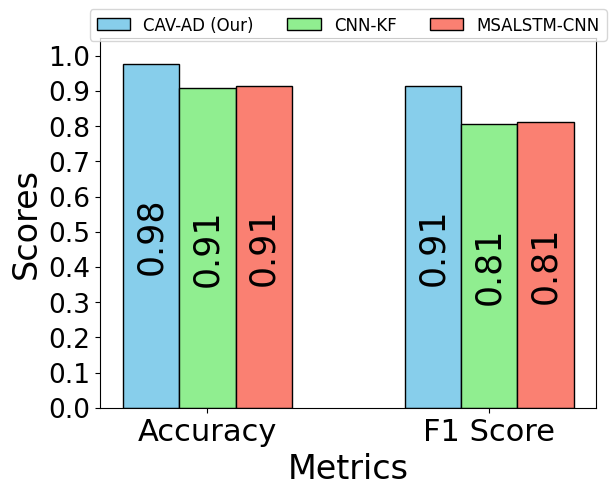}
         \caption{Sensor\#2.}\label{fig:Result_Constant_Sensor2}
     \end{subfigure}  
     \begin{subfigure}[b]{0.325\textwidth}
         \centering
         \includegraphics[width=\textwidth]{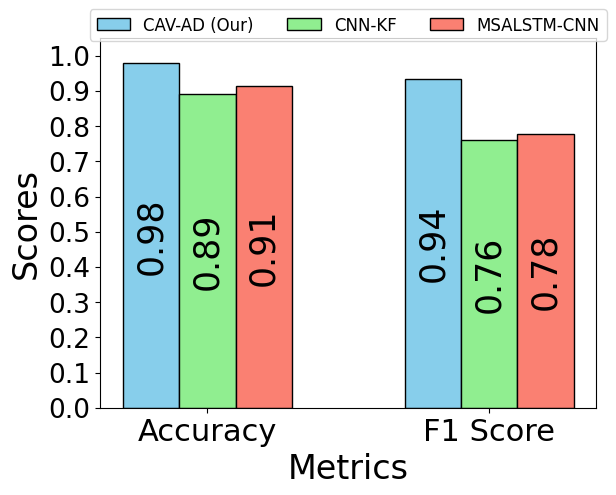}
         \caption{Sensor\#3.}\label{fig:Result_Constant_Sensor3}
     \end{subfigure}  
     \caption{Comparison of Accuracy and F1 score among CAV-AD, CNN-KF, and MSALSTM-CNN for constant anomaly detection.}
     \label{result:cons}
\end{figure*} 
%Evaluation
\section{Performance Evaluation}\label{sec:evaluation}
\subsection{Experimental Setup} 
{\bf Environment.} We employ Python 3.12.1 along with PyTorch packages to validate the effectiveness of CAV-AD. The machine specifications consist of a 64-bit operating system, 32 GB memory, and an AMD Ryzen 7 7700X 8-Core Processor paired with an NVIDIA GeForce RTX 3070 graphics card.

{\bf Dataset, Training Parameters, and Evaluation Metrics.} 
We utilize the safety pilot model deployment (SPMD) dataset \cite{bezzina2014safety} to evaluate our CAV-AD approach. This dataset comprises real recorded high-frequency data collected every 100 milliseconds over a two-year period for more than 2,500 vehicles that form a CAV network $\mathcal{C}$. The dataset includes both inter- and intra-vehicle communication, as well as vehicle-to-infrastructure communications. Each vehicle $c_i$ is equipped with three sensors $\mathcal{S}_{c_i}=\{s_1,s_2,s_3\}$: $s_1$ measures in-vehicle speed, $s_2$ measures GPS speed, and $s_3$ measures in-vehicle acceleration, capturing vehicle acceleration data. The trip duration for each vehicle $C_I$ is 2,980 seconds. To simulate anomaly scenarios, we use the anomalies generated in \cite{van2019real}, which introduce anomalies at a rate of 5\% as the original SPMD dataset does not contain any anomalies.

For the training of the O-OS-CNN model, we utilize the Adam optimizer with a learning rate of 0.0001 and a batch size of 20. We employ the CrossEntropyLoss loss function for optimization \cite{zhang2018generalized}.

The performance of CAV-AD is evaluated using the following metrics: accuracy (ACC), precision (Prec), F1 score, and confusion matrix.

{\bf Baselines.} We evaluate our approach, CAV-AD, against various existing methods from the literature, such as CNN-KF \cite{van2019real} and MSALSTM-CNN \cite{9210741}, using the same SPMD dataset \cite{bezzina2014safety} as they did.

\subsection{Comparison with SOTA}

 %As discussed in section 3.2, the train dataset is comprised of two anomaly-instant and constant whereas the test dataset has four types of anomaly. Thus our proposed model is trained on two anomaly types and  detection performance is evaluated on a test set comprising of multiple anomaly. The goal of the model is to detect instant anomaly or constant anomaly separately from a mixer of anomaly data. 
\begin{figure*}
     \centering
     \begin{subfigure}[b]{0.33\textwidth}
         \centering
         \includegraphics[width=\textwidth]{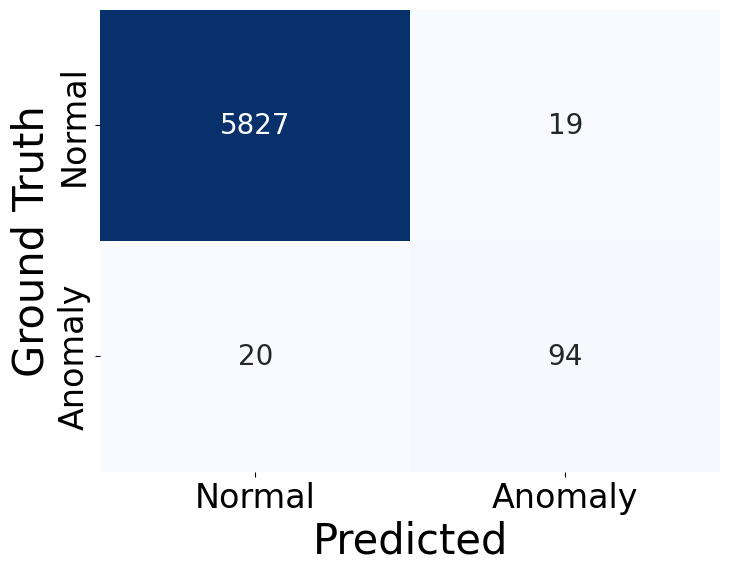}
        \caption{Sensor\#1.}\label{fig:Result_Instant_Sensor1_conf}
     \end{subfigure}  
     \begin{subfigure}[b]{0.325\textwidth}
         \centering
         \includegraphics[width=\textwidth]{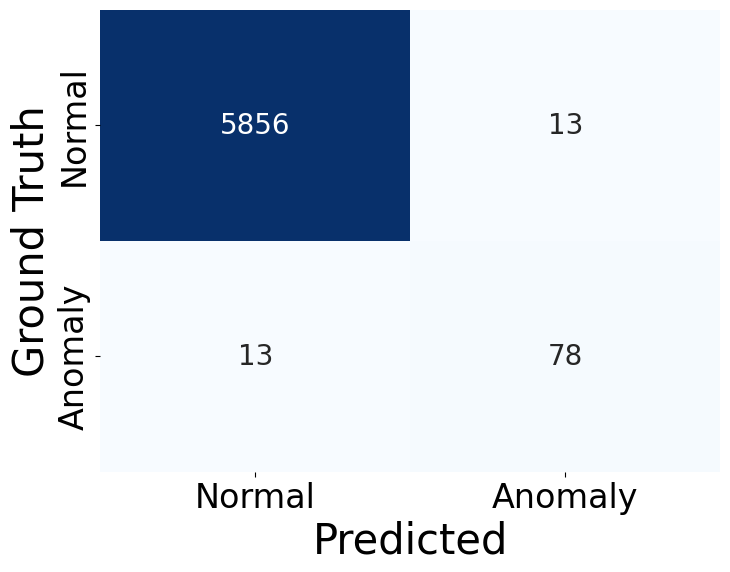}
         \caption{Sensor\#2.}\label{fig:Result_Instant_Sensor2_conf}
     \end{subfigure}  
     \begin{subfigure}[b]{0.325\textwidth}
         \centering
         \includegraphics[width=\textwidth]{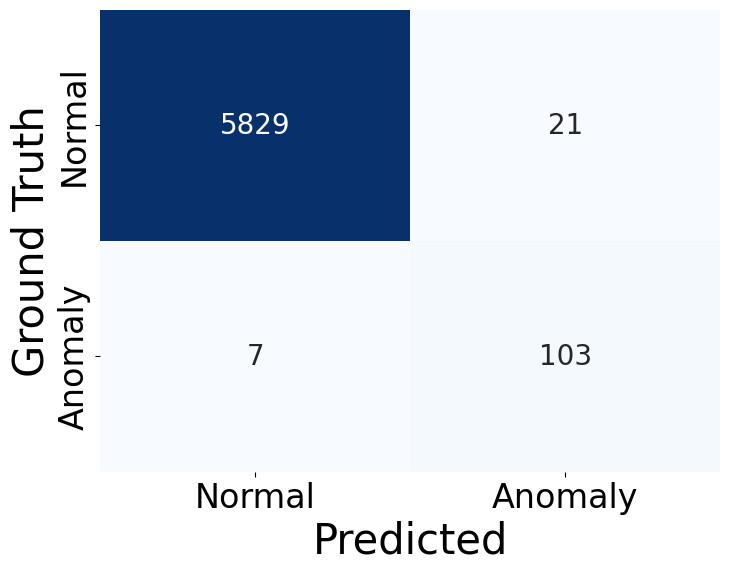}
         \caption{Sensor\#3.}\label{fig:Result_Instant_Sensor3_conf}
     \end{subfigure}  
     \caption{Confusion matrix  instant anomaly detection.}
     \label{result:instant_conf}
\end{figure*}

%Confusion Matrix-Constant
\begin{figure*}
     \centering
     \begin{subfigure}[b]{0.325\textwidth}
         \centering
         \includegraphics[width=\textwidth]{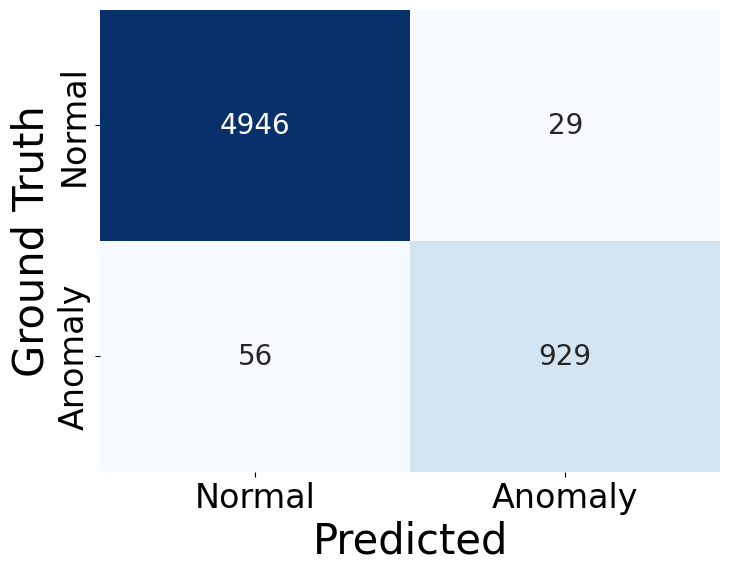}
         \caption{Sensor\#1.}\label{fig:Result_Constant_Sensor1_conf}
     \end{subfigure}  
     \begin{subfigure}[b]{0.325\textwidth}
         \centering
         \includegraphics[width=\textwidth]{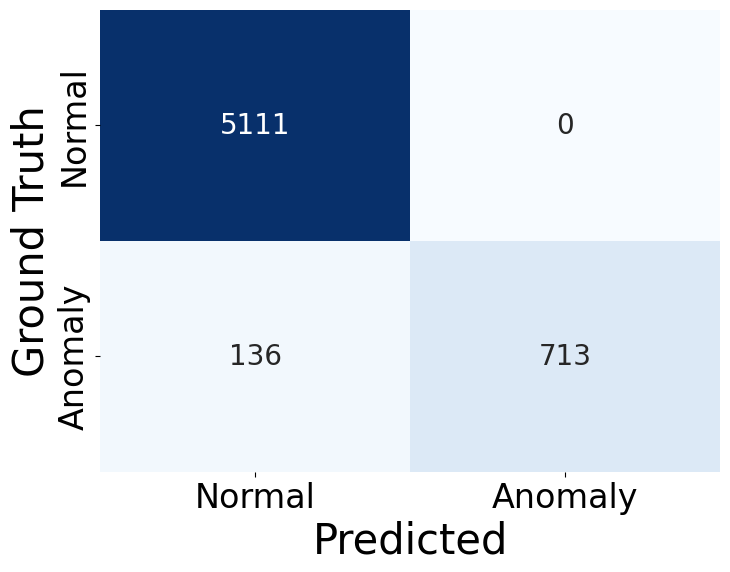}
         \caption{Sensor\#2.}\label{fig:Result_Constant_Sensor2_conf}
     \end{subfigure}  
     \begin{subfigure}[b]{0.325\textwidth}
         \centering
         \includegraphics[width=\textwidth]{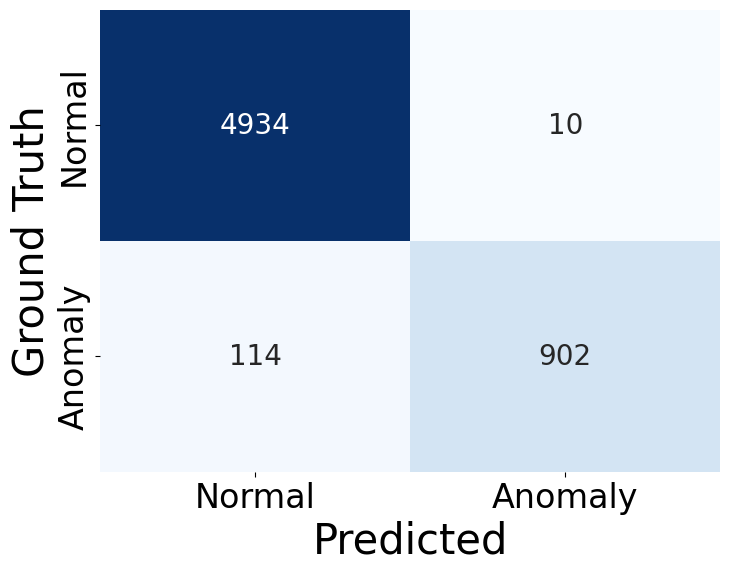}
         \caption{Sensor\#3.}\label{fig:Result_Constant_Sensor3_conf}
     \end{subfigure}  
     \caption{Confusion matrix constant anomaly detection.}
     \label{result:constant_conf}
\end{figure*} 

{\bf Performance of Detecting anomalies data.} We evaluate the performance of our approach, CAV-AD, against our baselines in detecting unknown anomalies, testing it over four types of anomalies as discussed in \sref{sec:threat model}, focusing on detecting instant and constant attacks.

Table~\ref{table:instant} displays CAV-AD's detection performance for instant anomalies, while Table~\ref{table:constant} presents performance for constant anomalies according to our evaluation metrics. In Table~\ref{table:instant}, $\mathcal{N}(0,0.01)$ defines a Gaussian variable \cite{goodman1963statistical} with mean and variance of zero and 0.01, respectively. This variable is amplified by 1000 and added to the base value of sensor measurements for one epoch (i.e., 100 ms). In Table~\ref{table:constant}, $\mathcal{U}(0,5)$ refers to a uniform distribution between 0 and 5, and $d=10s$  is the duration of the anomaly. For both types of attacks, our findings show that the accuracy for all three sensors is above 97\%. The F1 score for detecting instant anomalies from $s_1$, $s_2$, and $s_3$ is 83.4\%, 88.7\%, and 88\%, respectively, which is less than 90\%. On the other hand, the F1 score for all three sensors in constant anomalies is above 90\%. It is also evident that the precision follows a similar trend to the F1 score. The reason for this difference is that instant anomalies have no duration, unlike constant anomalies, and are more challenging to identify due to random shifts in magnitude at random time instances. However, this performance supports the strong generalization of the model in detecting these two significant anomaly types.

In \fref{result:instant}, we present the comparison results of CAV-AD against the baselines for the instant anomaly, showing that CAV-AD outperforms the current CNN-KF \cite{van2019real} and MSALSTM-CNN \cite{9210741} approaches in terms of accuracy and F1 score. While our model achieves the same accuracy of 99\% across all three sensors, it achieves an F1 score of 83\% for $s_1$, 88\% for $s_2$, and 88\% for $s_3$, representing an improvement of 5-27\% on F1 score compared to the baselines. These results show a significant improvement in F1 scores, underscoring the effectiveness of CAV-AD over its counterparts. This improvement can be attributed to the diverse range of kernel sizes utilized in training the O-OS-CNN model.

In \fref{result:cons}, we extend the evaluation to identify constant anomalies from mixed anomaly data, achieving an accuracy of 98\% across the three sensors. The F1 score for $s_1$, $s_2$, and $s_3$ is 96\%, 91\%, and 94\%, respectively. These results further validate the robustness and effectiveness of CAV-AD in detecting anomalies across various scenarios.

Lastly, in \fref{result:instant_conf} and \fref{result:constant_conf}, we present the confusion matrices for instant and constant anomalies, respectively, to illustrate how well our model distinguishes between normal and anomaly data. It is evident that the number of true positive instances (anomaly) is significantly higher in the case of constant anomalies compared to instant anomalies, indicating a class imbalance problem. As there are more anomaly data instances in constant anomalies than in instant anomalies, the model identifies more instances as true positives in constant anomalies. This observation suggests that providing more labeled data of the anomaly class to the proposed model would enhance anomaly detection accuracy. Overall, our model successfully detects a high number of anomaly instances with minimal misclassifications.

%Detection of Attacked Sensor
{\bf Evaluation of Malicious Sensor(s) Detection.}  Pertaining to Algorithm \ref{algo:detection} discussed in \sref{sec:methodology} for detecting and identifying malicious sensors, a significant aspect of our CAV-AD framework,we evaluate the effectiveness of this algorithm in detecting instant anomalies.

\begin{figure*}
    \centering
    \begin{subfigure}[b]{0.325\textwidth}
        \centering
        \includegraphics[width=\textwidth]{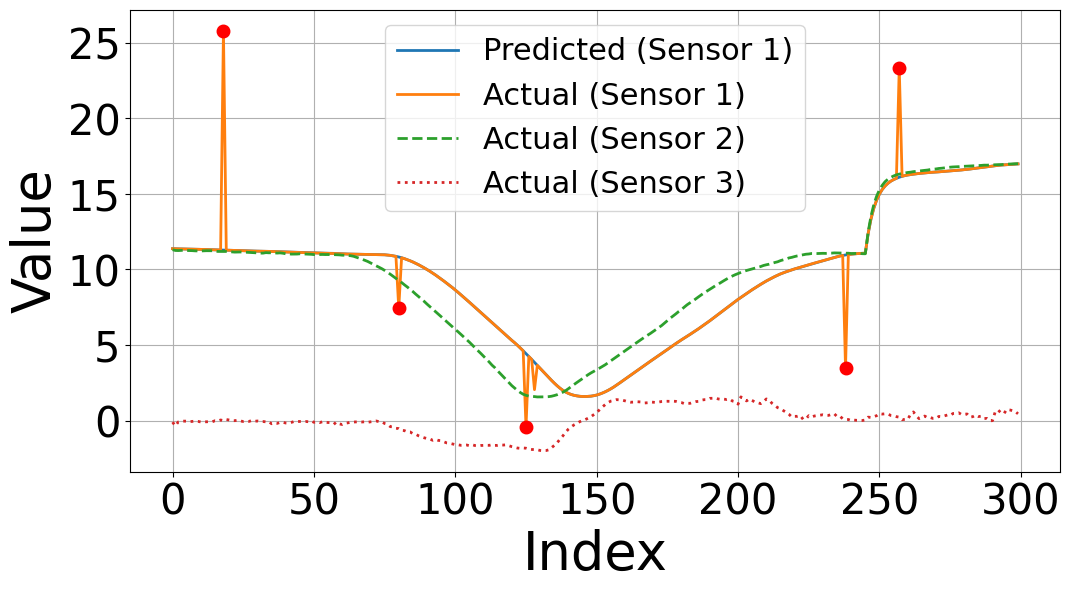}
        \caption{Sensor\#1.}\label{fig:KF_Sensor1}
    \end{subfigure}
    \hfill
    \begin{subfigure}[b]{0.325\textwidth}
        \centering
        \includegraphics[width=\textwidth]{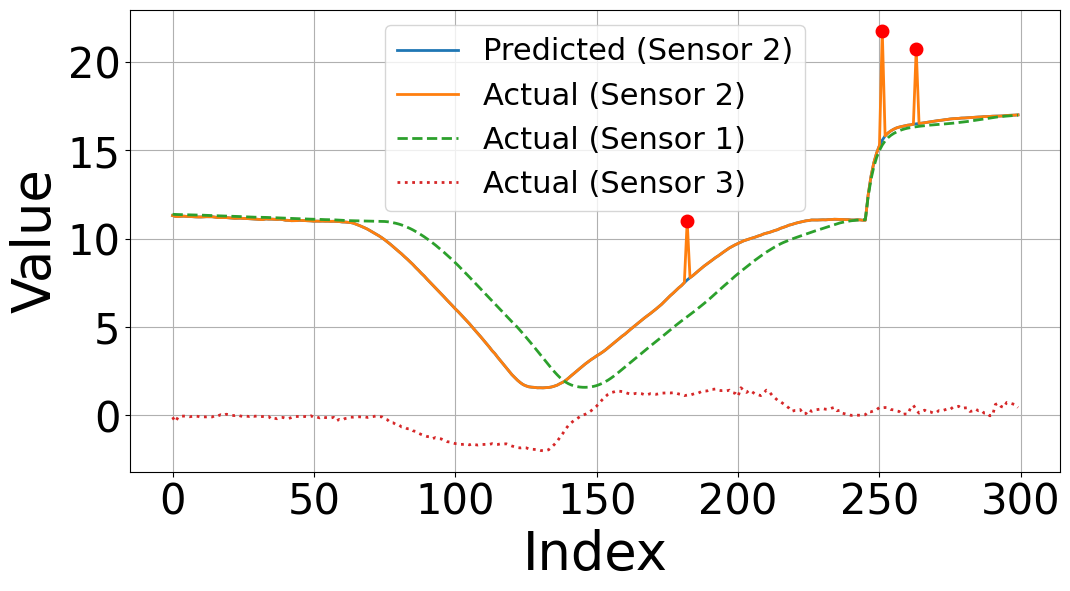}
        \caption{Sensor\#2.}\label{fig:KF_Sensor2}
    \end{subfigure}
    %\vspace{0.5cm} % Adjust the vertical space between rows
    \begin{subfigure}[b]{0.325\textwidth}
        \centering
        \includegraphics[width=\textwidth]{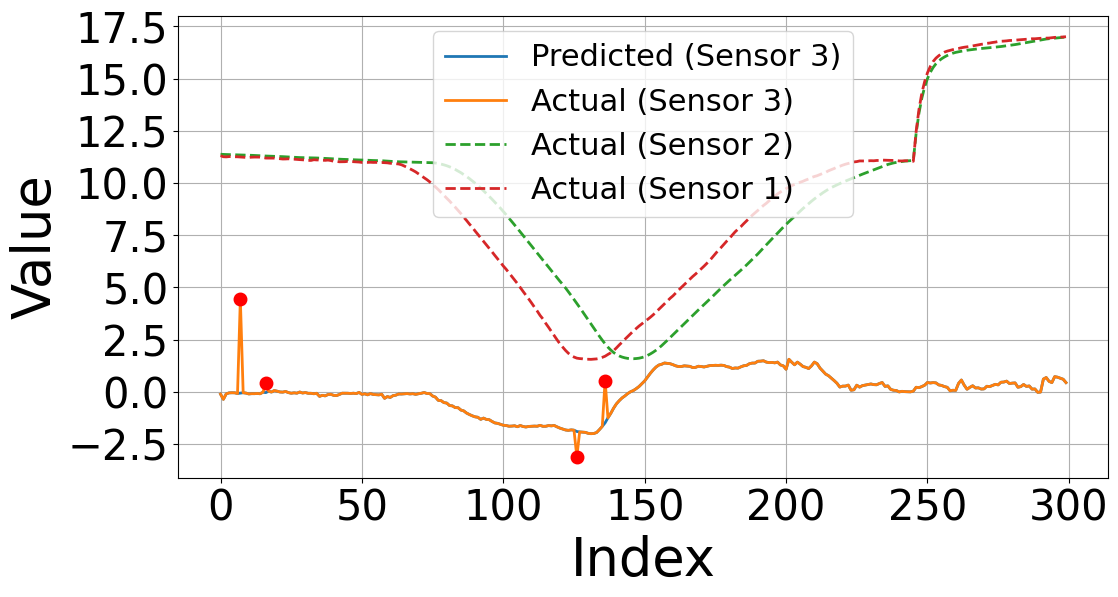}
        \caption{Sensor\#3.}\label{fig:KF_Sensor3}
    \end{subfigure}
    \caption{Instant anomaly detected using KF for various sensors.}
    \label{fig:kf}
\end{figure*}

In \fref{fig:KF_Sensor1}, the detection of anomalies from $s_1$ is depicted. The predicted values by the KF are shown in blue, closely in line with the actual values represented by the orange line. When the incoming anomaly reading deviates from the normal reading, the KF identifies it as abnormal (marked by red circles). This pattern is consistent for $s_2$ and $s_3$ as illustrated in \fref{fig:KF_Sensor2} and \fref{fig:KF_Sensor3}, respectively.

\begin{figure*}
    \centering
    \begin{subfigure}[b]{0.325\textwidth}
        \centering
        \includegraphics[width=\textwidth]{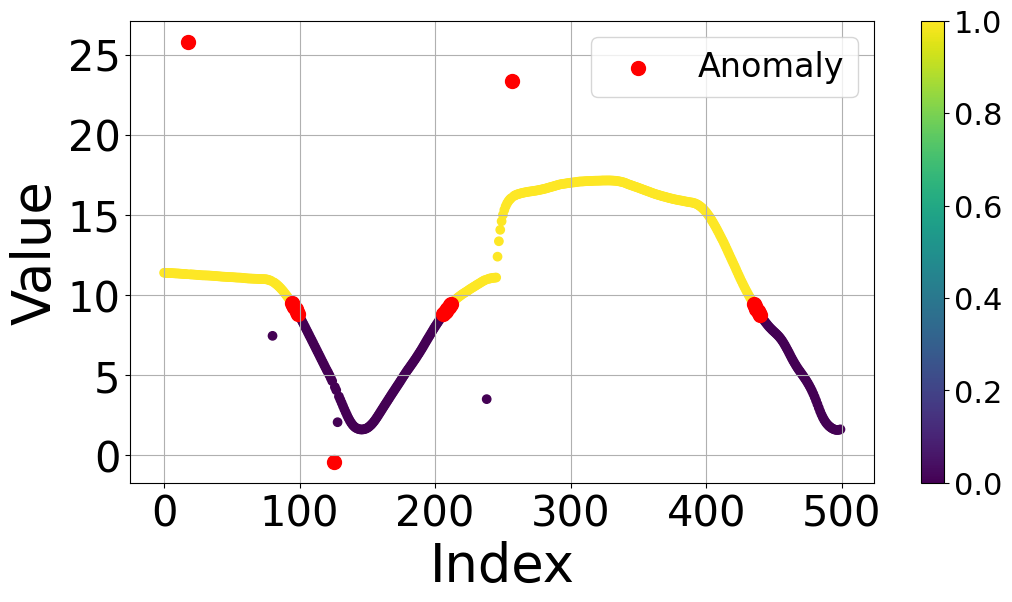}
        \caption{Sensor\#1.}\label{fig:GMM_Sensor1}
    \end{subfigure}
    \begin{subfigure}[b]{0.325\textwidth}
        \centering
        \includegraphics[width=\textwidth]{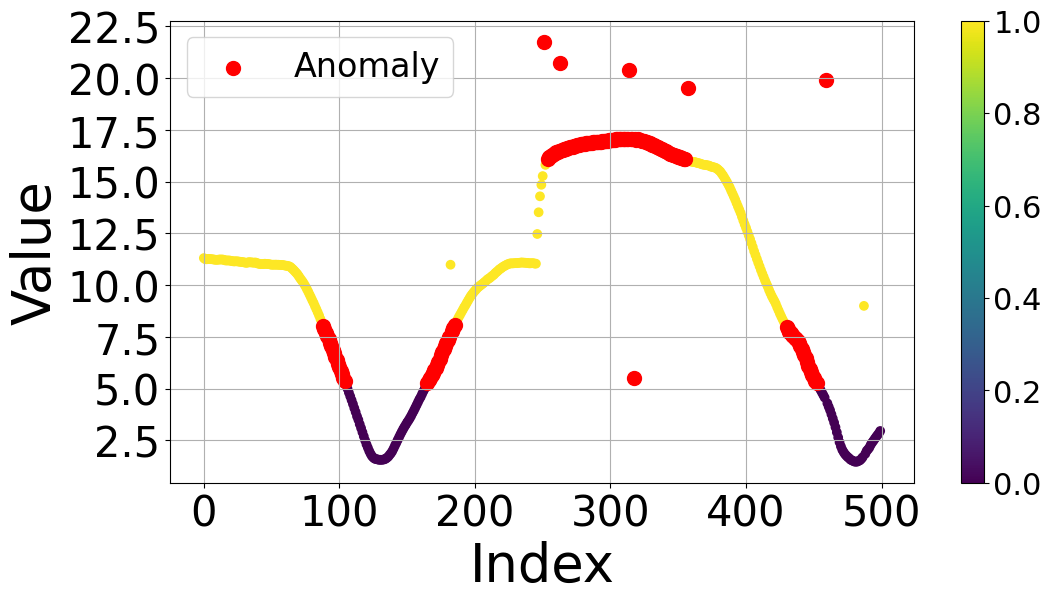}
        \caption{Sensor\#2.}\label{fig:GMM_Sensor2}
    \end{subfigure}
    \begin{subfigure}[b]{0.325\textwidth}
        \centering
        \includegraphics[width=\textwidth]{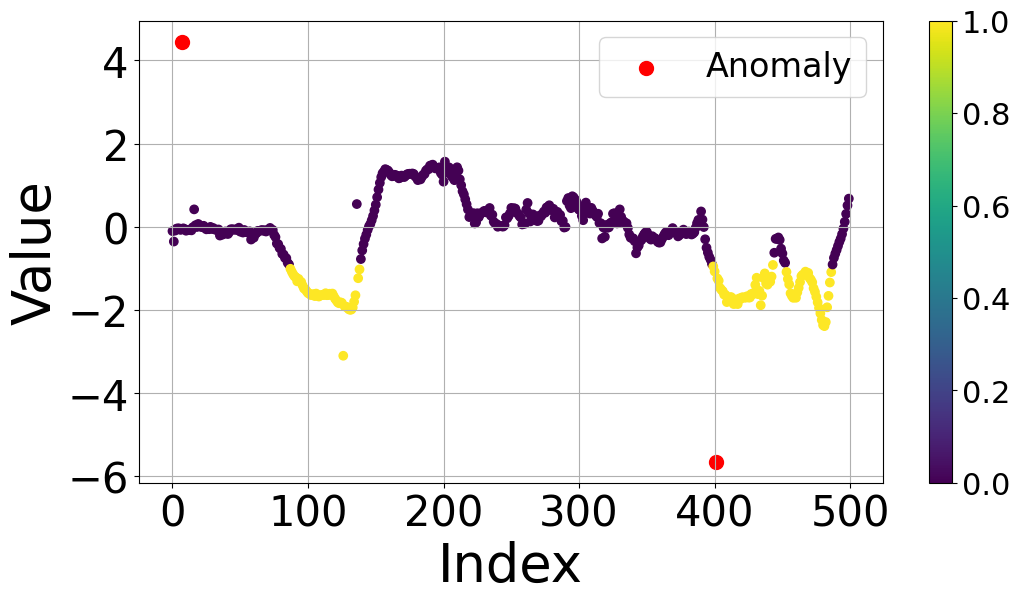}
        \caption{Sensor\#3.}\label{fig:GMM_Sensor3}
    \end{subfigure}
    \caption{Instant anomaly detected using GMM for various sensors.}  \label{fig:GMM_Sensor}
\end{figure*}

To further demonstrate the effectiveness of KF over other methods, we compare the results with GMM in \fref{fig:GMM_Sensor}. Here, GMM creates two separate clusters represented in dark purple and yellow, with the red cluttered dots indicating anomaly data. In addition, the red dots at the junction of two clusters are normal sensor values, indicating that GMM with O-OS-CNN model fails to accurately detect anomaly data occurring in a specific sensor. In contrast, KF with O-OS-CNN successfully achieves this detection. This phenomenon is observed consistently across all three sensors, highlighting the effectiveness of KF over GMM.

\begin{figure*}
     \centering
     \begin{subfigure}[b]{0.325\textwidth}
         \centering
         \includegraphics[width=\textwidth]{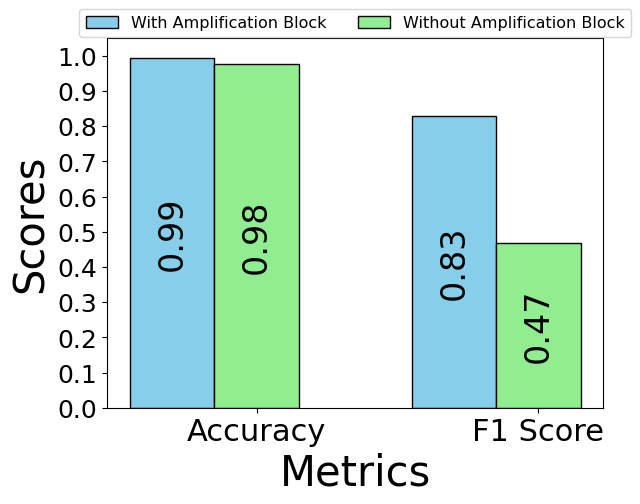}
         \caption{Sensor\#1.}\label{fig:Ablation_Instant_Sensor1}
     \end{subfigure}  
     \begin{subfigure}[b]{0.325\textwidth}
         \centering
         \includegraphics[width=\textwidth]{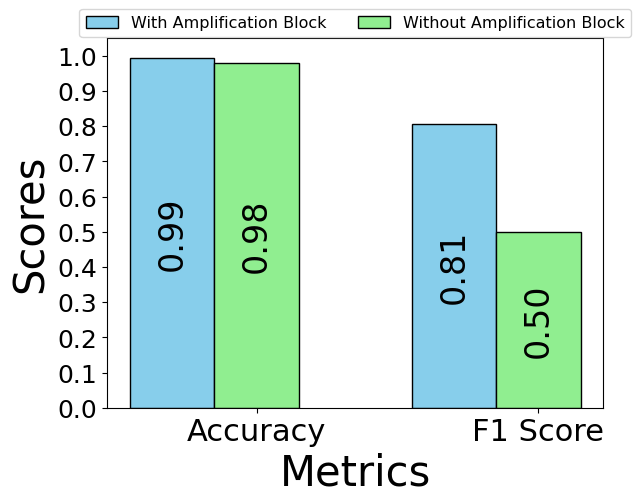}
         \caption{Sensor\#2.}\label{fig:Ablation_Instant_Sensor2}
     \end{subfigure}  
     \begin{subfigure}[b]{0.325\textwidth}
         \centering
         \includegraphics[width=\textwidth]{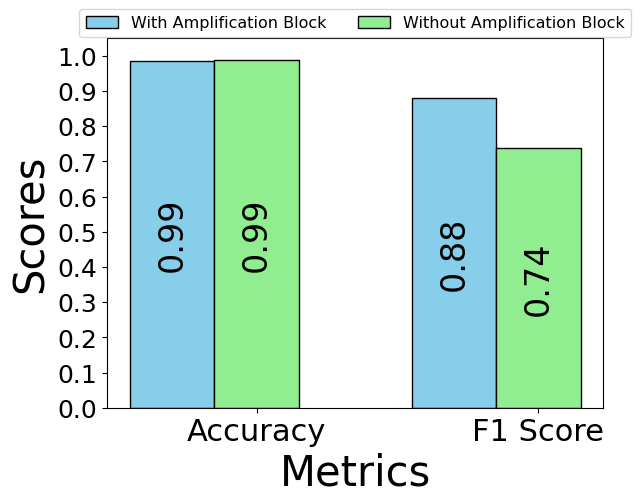}
         \caption{Sensor\#3.}\label{fig:Ablation_Instant_Sensor3}
     \end{subfigure}  
     \caption{Evaluation of the amplification block within the CAV-AD framework for instant anomaly detection.}
     \label{fig:ablation_instant}
\end{figure*}

\begin{figure*}
     \centering
     \begin{subfigure}[b]{0.325\textwidth}
         \centering
         \includegraphics[width=\textwidth]{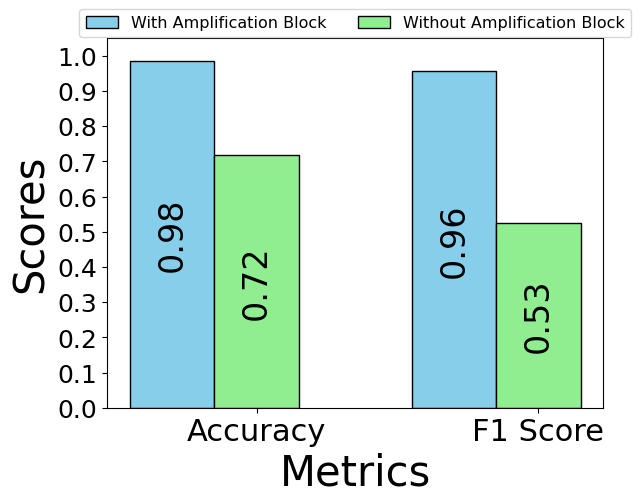}
         \caption{Sensor\#1.}\label{fig:Ablation_Constant_Sensor1}
     \end{subfigure}  
     \begin{subfigure}[b]{0.325\textwidth}
         \centering
         \includegraphics[width=\textwidth]{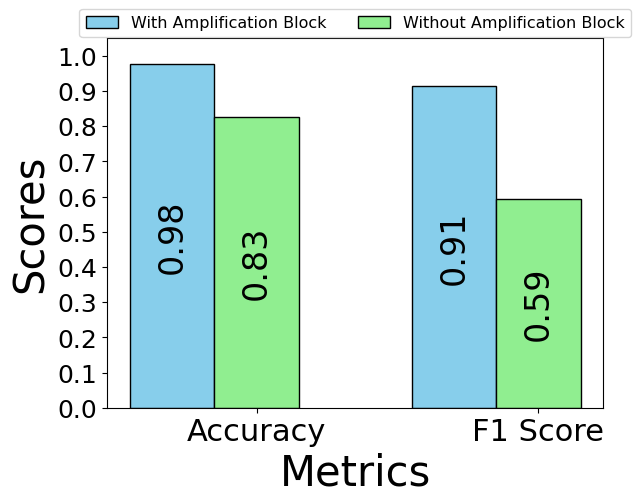}
         \caption{Sensor\#2.}\label{fig:Ablation_Constant_Sensor2}
     \end{subfigure}  
     \begin{subfigure}[b]{0.325\textwidth}
         \centering
         \includegraphics[width=\textwidth]{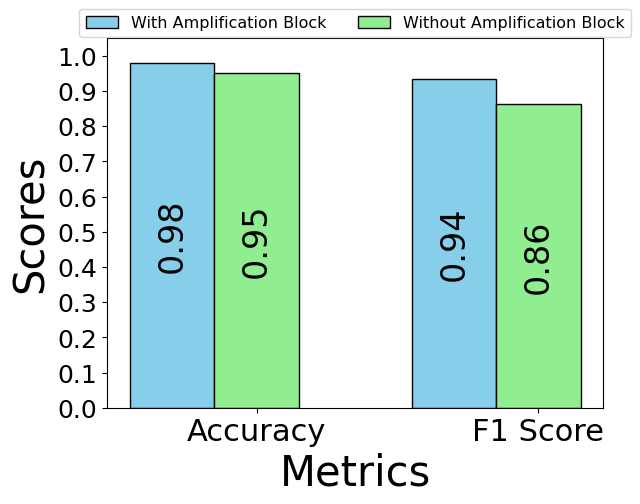}
         \caption{Sensor\#3.}\label{fig:Ablation_Constant_Sensor3}
     \end{subfigure}  
     \caption{Evaluation of the amplification block within the CAV-AD framework for the constant anomaly.}\label{fig:Ablation_Constant_Sensor}
\end{figure*}

%Ablation Study Result
{\bf Performance Evaluation of Amplification Block.}  In this section, we assess the significance of the amplification block within the CAV-AD framework through ablation experiments. To achieve this, we conduct separate experiments with and without the amplification block integrated into the CAV-AD framework. In \fref{fig:ablation_instant}, we present the accuracy and F1 score of all three sensors' detection performance for instant anomaly detection, both with and without the amplification block. The results indicate a notable enhancement in overall performance when the amplification block is included. Similar improvements are also observed for constant anomaly detection as shown in \fref{fig:Ablation_Constant_Sensor}. This is because higher amplitude anomalies are more pronounced and serve as better features for the model to extract. Thus, it is evident that our proposed CAV-AD framework has the capability to effectively detect real-world anomalies characterized by significant deviations from normal values.

\section{Conclusion and Future Work}
In this paper, we propose a novel framework named CAV-AD designed to detect anomalous readings of sensor data in CAV networks. The framework not only identifies anomalies in sensor data but also identifies sensors that have been compromised. Specifically, CAV-AD stands out against the SOTA by three components: {\em a novel O-OS-CNN model architecture} that extracts features from the input data by generating all possible kernel sizes of varying lengths, {\em an amplification block} that enhances the model's capability to distinguish anomaly data, and {\em the incorporation of KF with O-OS-CNN} for instant detection of malicious sensors. Our experimental results demonstrate significant improvements over the SOTA in detecting anomalies among a mixed type of anomaly by an order of magnitude using a real-world dataset for CAV networks, the SPMD dataset. CAV-AD achieves higher accuracy in detecting both instant and constant anomalies, with an accuracy above 90\% and a higher F1 score of 96\%, representing an improvement of 5-27\% compared to SOTA.

In future, to extend the capabilities of the CAV-AD framework, we will explore the detection of other types of anomalies. %We intend to reduce the computational overhead to deploy in real-world. 
Additionally, we plan to investigate AD in more complex vehicle environments for deployment in real-world scenarios. Furthermore, we intend to delve into the detection of more precise anomaly data, which presents a greater challenge. Determining the specific type of anomaly occurring could also aid in the development of real-time remedies.
\label{sec:conclusion}

\bibliographystyle{IEEEtran}
\renewcommand{\baselinestretch}{.9}
\small{
\bibliography{_references.bib}
}
\end{document}